# MedAL: Accurate and Robust Deep Active Learning for Medical Image Analysis

Asim Smailagic, Hae Young Noh, Aurélio Campilho, Pedro Costa, Devesh Walawalkar, Kartik Khandelwal,
Mostafa Mirshekari, Jonathon Fagert, Adrian Galdran, Susu Xu

*Abstract*—Deep learning models have been successfully used in medical image analysis problems but they require a large amount of labeled images to obtain good performance. However, such large labeled datasets are costly to acquire. Active learning techniques can be used to minimize the number of required training labels while maximizing the model's performance. In this work, we propose a novel sampling method that queries the unlabeled examples that maximize the average distance to all training set examples in a learned feature space. We then extend our sampling method to define a better initial training set, without the need for a trained model, by using Oriented FAST and Rotated BRIEF (ORB) feature descriptors. We validate MedAL on 3 medical image datasets and show that our method is robust to different dataset properties. MedAL is also efficient, achieving 80% accuracy on the task of Diabetic Retinopathy detection using only 425 labeled images, corresponding to a 32% reduction in the number of required labeled examples compared to the standard uncertainty sampling technique, and a 40% reduction compared to random sampling.

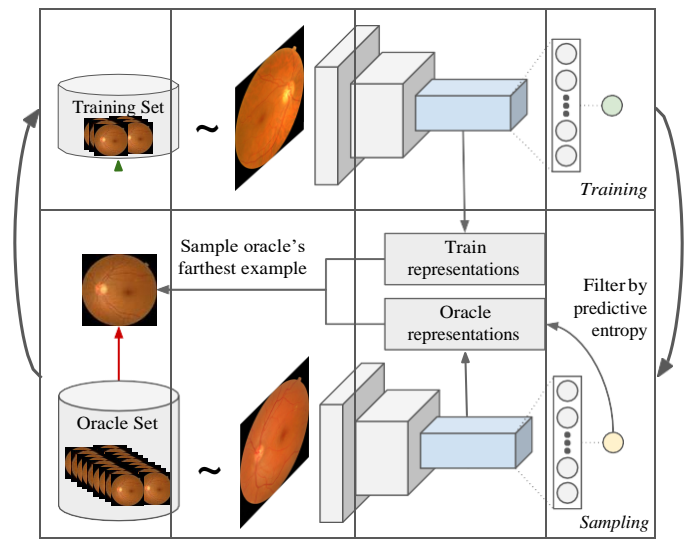

Fig. 1. **Proposed Active Learning pipeline.** We start by training a model and then use it to query examples from an unlabeled dataset that are then added to the training set. In this work we propose a novel query function that is better suited for Deep Learning models. We use the DL model to extract features from both the oracle and training set examples, and then we filter out the oracle examples that have low predictive entropy. Finally, we select the oracle example that is on average the most distant in feature space to all training examples.

## I. INTRODUCTION

Active Learning (AL) techniques aim at reducing the amount of manual example annotation required to train large-scale machine learning models by selective sampling of an unlabeled dataset. Therefore, a natural field of application for AL is medical image analysis, where the manual labeling work is often expensive, time-consuming, and laborious, requiring the intervention of medical experts. The relevance of AL in medical imaging applications has steadily grown on pair with the progressive success in the field of deep learning models [1], which are greatly demanding in terms of annotated data.

Many different medical image modalities have benefited from AL in recent years. In the field of histopathological image analysis, AL-enhanced models have been employed to adapt existing computational diagnosis techniques to highly-imbalanced scenarios [2], or for cell nucleus segmentation [3].

In CT scans analysis, optimal training set construction for image segmentation tasks has been explored via AL-based sampling, while in MRI analysis, interactive segmentation models that can query the user about the most relevant input areas in an image have also been proposed [4]. Computer-Aided Diagnosis of digital mammograms [5] has also been addressed via AL applied to decide which of different views of a breast exam is most useful for a mass diagnosis system, and a cost-effective sampling method for minimizing a skin lesion segmentation model uncertainty (as measured by Monte Carlo Sampling) has been recently developed [6]. The combination of Multiple Instance Learning approaches with AL for tuberculosis detection on chest radiographies has also been recently explored in [7], where an instance selection framework is proposed to supply meaningful image regions to a classifier instead of an entire individual instance.

Traditional AL techniques build mainly upon different sampling schemes: Uncertainty Sampling, Query By Committee, Expected Model Change, and Expected error reduction. These approaches suffer from different limitations, as will be reviewed in the next section. Generally speaking, computational cost and sensitivity to irregular feature scaling are their major drawbacks. In order to overcome these limitations, we introduce MedAL, a novel AL approach which leverages a combination of predictive entropy based uncertainty sampling and a distance function on a learned feature space to optimize the selection of unlabeled samples, as described in Figure 1. This combined approach ensures that each training iteration selects the images providing the most information about the entire dataset for the model being trained, thereby minimizing the total number of images required for training. MedAL overcomes the limitations of the traditional approaches by efficiently selecting only the images that provide the most information about the overall data distribution, effectively reducing computation cost and increasing both speed and accuracy. Our approach is validated by conducting experiments on a series of medical image diagnosis tasks and modalities, namely: diabetic retinopathy detection from retinal fundus images, breast cancer grading from histopathological images, and melanoma diagnosis from skin images.

The **main contributions** in this paper are therefore:
- *Novelty:* we present a novel AL sampling method that queries the unlabeled examples that maximize the average distance to all training set examples in a learned feature space.
- *Better Initialization:* we extend our sampling method to define a better initial training set, without a trained model, by using standard feature extraction and description methods.
- *Efficient:* our method achieves better results with fewer labeled examples than competing methods.
- *Robust:* we test our method on binary and multi-class classification problems using balanced and unbalanced datasets.

## II. RELATED WORK

Every AL scenario involves determining the information contained in unlabeled instances, which can be sampled from a given distribution. There have been many proposed ways of formulating such sampling/query strategies in the literature. Among the most popular AL techniques we can find: Query By Committee (QBC), Expected Error Reduction (EER), Expected Model Change (EMC), and Uncertainty Sampling (US). Let us quickly review the main principles and limitations underlying each of these approaches.

*1) Query By Committee:* QBC is an effective approach to selective sampling in which a committee of $n$ student models is trained on the same dataset. The next queried data sample is selected from this dataset based on maximizing the disagreement among the predictions generated from all models. The main idea behind QBC approach is thereby to minimize the version space, *i.e.* the set of hypotheses that are consistent with the labeled training data. QBC is computationally inefficient as it requires training multiple models.

*2) Expected Error Reduction:* EER is based on estimating the degree of reduction in the "future error" when a new instance in the dataset is labeled, and this information becomes available for training. Since there is no knowledge on the labels of the dataset, EER estimates the average-case criterion of potential loss instead [8]. In this context, the instance with minimal risk (expected future error) is the one to be queried next. This technique is computationally costly as it requires to estimate the future error for each query in the oracle set and then re-train the model for each possible query labeling.

*3) Expected Model Change:* EMC is an approach for querying an example which would cause a significant change in the model, where its label is to be known [9]. It has two major drawbacks. First, it does not perform adequately when the features are not properly scaled. Second, it can be computationally expensive if both the feature space and set of unlabeled data are very large. Some extensions of EMC have been proposed to deal with the original EMC inefficiency, like Variance Reduction [10], where examples that reduce the model variance the most are selected. Unfortunately, EMC methods are still empirically much slower than simpler query strategies like Uncertainty Sampling.

*4) Uncertainty Sampling:* US is another popular sampling technique [11]. In this method an active learner queries the instances about which it is least certain. The degree of uncertainty can be calculated by various methods such as the entropy measure, which is a good estimate of the degree of randomness. In US, the least certain instance typically lies on the classification boundary, but this does not mean it needs to be "representative" of other instances in the data distribution. Hence, just knowing its label is unlikely to significantly improve accuracy on the dataset as a whole.

*5) Other AL approaches:* Other different techniques beyond the ones outlined above have also proposed in literature trying to better capture the representation of the data. For example, clustering has been proposed as a way of accelerating and improving the sampling process in [12], and a hierarchical strategy similar in spirit has been proposed in [13].

All the techniques are generic, and unlike our proposed approach do not leverage the powerful feature extraction capabilities of modern neural networks to efficiently query informative images. In the next section, we will give a detailed description of our method.

## III. PROPOSED METHOD

Most existing AL techniques have been developed for classical machine learning methods with much lower learning capacity than the most recent *Deep Neural Network* (DNN) architectures. These DNNs have been shown to be even capable of fitting a random labeling of the training data [14], and they do so with high confidence (low uncertainty). This ability reduces the effectiveness of traditional AL methods that rely on the degree of uncertainty in the prediction of the model in order to sample new useful examples to the training set.

In this section we present a method that is better suited to be used with DNNs. We start by describing our novel sampling method and then we show how to combine it with a DNN to perform AL on image recognition tasks. We also describe a mechanism to build a representative initial dataset when the training process has not still started.

### A. Sampling Based on Distance between Data Descriptors

Let $D_{train}$ be the training set and $D_{oracle}$ be the unlabeled dataset from which new examples can be sampled and labeled on-demand. In an AL setting we train a model using $D_{train}$ and then use the trained model to analyze $D_{oracle}$ in order to find the most informative example $x^*$. Then, $x^*$ is labeled and added to $D_{train}$, and the training of the model starts over. This process is iterated with the main objective of minimizing the labeling effort while maximizing the model's performance.

When analyzing $D_{oracle}$, for each example $x$ we can consider the prediction $y$ generated by the model in order to decide about how informative $x$ is. This is possible since $y$ is usually constrained in a well known range (*i.e.* between 0 and 1 in a binary classification problem), which makes it possible to directly evaluate the uncertainty of the model on $x$.

In this work, we depart from the standard practice of only using the model's prediction to decide which example should be sampled from $D_{oracle}$. Instead, we propose to use rich data descriptors $f(x) \in R^n$, which are likely to be more informative about the usefulness of $x$. Moreover, for comparing the degree of relevant new information provided by $x$, we compare an example's representation with all samples in $D_{train}$ by means of a distance function $d$:

$$s(x) = \frac{1}{N} \sum_{i=1}^{N} d(f(x_i), f(x)), \quad (1)$$

where $x_i \in D_{train}$. In following sections, we will describe in detail the extraction of data representation $f(x)$ and provide an analysis on different possibilities for selecting $d$.

Instead of computing $s(x)$ for every $x \in D_{oracle}$, we first compute the predictive entropy on each element, and select the subset containing the $M$ samples with maximum entropy. We then calculate $s(x)$ for every element in this subset, and finally sample the example that maximizes $s$ within it:

$$x^* = \arg\max_{x \in D_{oracle}} s(x). \quad (2)$$

Next, $x^*$ is removed from $D_{oracle}$, labeled, and included in $D_{train}$. The model is iteratively retrained from scratch until performance on $D_{train}$ stops improving, and then used to sample a new example from $D_{oracle}$.

### B. Deep Representations as Data Descriptors

In order to extract powerful representations from the data and simultaneously solve an image classification problem, we employ a Convolutional Neural Network (CNN). This is meaningful, since it is known that the representation space learnt by CNNs contains semantic meaning, *e.g.* nearby elements in this space tend to be visually similar [15, 16]. Conversely, elements that are far away are typically different from each other. These properties are greatly relevant in order for Eq. (1) to perform as intended.

We therefore define the output of an intermediate CNN layer as the description extraction function $f$, as shown in Fig. 1. As the training progresses, the model will extract better representations that will in turn lead to both sampling more informative examples from $D_{oracle}$ and improving the final classification accuracy.

It is worth noting that in an AL context, the goal is to use the model that is being trained to sample new examples that can reduce its own uncertainty. Therefore, the feature extraction function $f(x)$ evolves while training the model. Otherwise, data descriptors could be computed off-line, and the order in which elements from $D_{oracle}$ are added to $D_{train}$ could be decided from the start, resulting into a static sampling strategy.

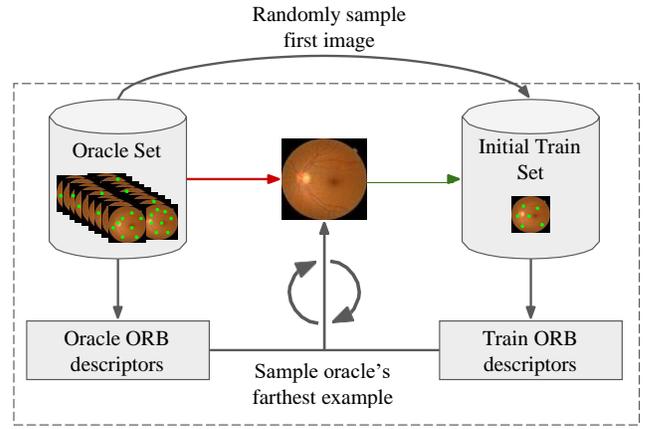

Fig. 2. **Initial Training Set Construction.** We start by randomly sampling an image from the oracle into the training set. Then, ORB descriptors are obtained for all oracle and train set images. The oracle image with farthest average distance to all training set images is added to the training set. This process is repeated until we get to the desired initial dataset size.

### C. Initial Training Set Construction

Conventional AL techniques typically form an initial training dataset by randomly extracting samples from the available dataset. Even though random sampling can theoretically capture the empirical data distribution, this is only guaranteed when a large number of samples is considered. However, when only a small initial dataset is available, random sampling may not be an optimal initialization strategy, as it can lead to sampling similar images together (*i.e.* data samples which provide almost the same information for the model).

In this work, we use the extraction function $f(x)$ in eq. (1) to form the initial dataset. There are, however, two obstacles that can render this strategy meaningless: $D_{train}$ is empty ($N = 0$); and 2) the model does not start from a trained state and, therefore, it will extract random representations.

In order to overcome the first problem, we randomly sample one image from $D_{oracle}$ to form the initial $D_{train}$. The second problem can be solved with two different approaches: using a pre-trained DNN to generate descriptors, or applying well-established extraction and description techniques from the computer vision field. In this work we use ORB descriptors [17], to account for a situation in which no pre-trained network is available. This approach is particularly well-suited for medical image analysis applications as these descriptors have been shown to perform well in the task of Diabetic Retinopathy detection [18].

For each image, ORB keypoints are detected and described into 256-bit vectors. These vectors are then average-pooled into a single image descriptor vector that can be used in Eq. (1). We can then add images to $D_{train}$ as previously explained by iteratively applying Eq. (1) and Eq. (2). Therefore, we do not use information about the dataset label distribution to build $D_{train}$, but we still manage to avoid adding similar images to the initial training set. This strategy is illustrated in Fig. 2.

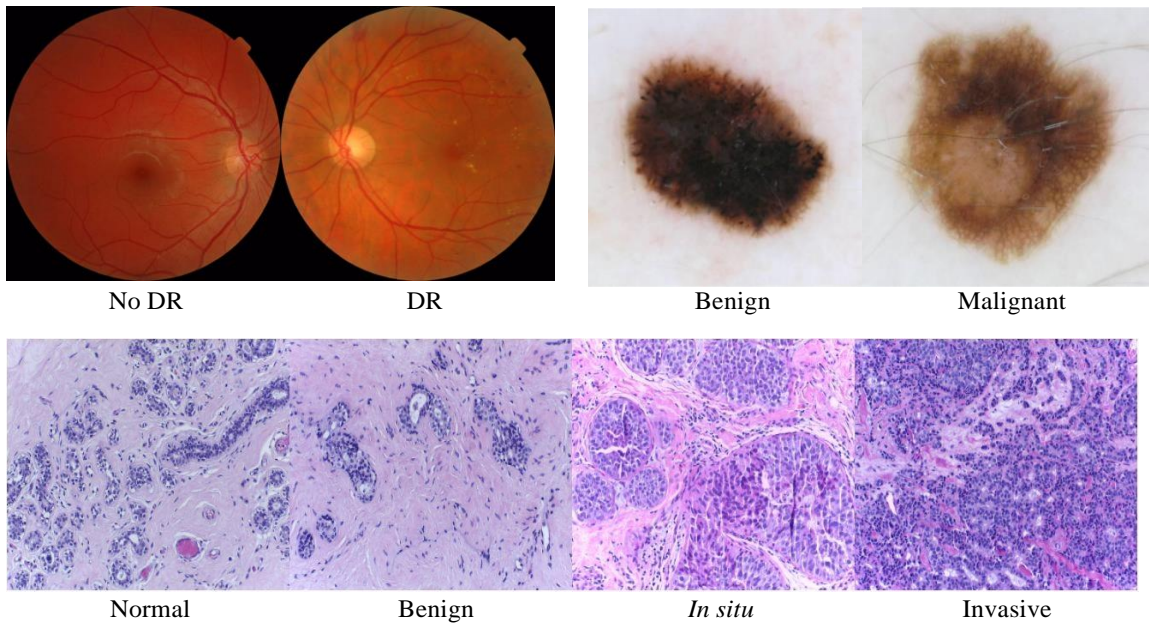

Fig. 3. **Examples from all three datasets evaluated in this work.** From left to right, top to bottom: the images belong to Messidor, Skin Cancer and Breast Cancer datasets. The class labels are shown bellow each of the images.

TABLE I
DATASET IMPLEMENTATION DETAILS. INFORMATION ABOUT EACH DATASET SIZE AND THE NUMBER OF IMAGES THAT ARE ADDED AT EACH AL ITERATION.

| Parameters | Datasets | | |
|---|---|---|---|
| | MESSIDOR | Breast Cancer | Skin Cancer |
| (train + oracle) size | 768 | 320 | 700 |
| Initial training set size | 100 | 30 | 100 |
| Ims added per iteration | 20 | 5 | 10 |
| $M$ | 50 | 30 | 50 |

TABLE II
OUR INITIAL DATASET FORMATION LEADS TO BETTER GENERALIZATION. TEST SET ACCURACY AFTER TRAINING THE MODEL UNTIL 100% ACCURACY ON AN INITIAL DATASET SAMPLED USING OUR METHOD AND RANDOM SAMPLING.

| Dataset | Random sampling | ORB Descriptor sampling |
|---|---|---|
| MESSIDOR | 57.89% | **65.23%** |
| Breast Cancer | 68.43% | **74.93%** |
| Skin Cancer | 41.37% | **46.35%** |

## IV. EXPERIMENTS

In order to validate the accuracy and robustness of our proposed approach, we performed experiments on three medical datasets. We first introduce the datasets (in Section IV-A). Next, we evaluate a set of common distance metrics to find the best distance function for our approach (in Section IV-C). Then, we present and analyze the effect of utilizing ORB descriptors (in Section IV-D) and finally we evaluate our sampling method on three different medical imaging datasets (in Section IV-E).

### A. Dataset Description

By reducing the labeling requirement, active learning is suitable for identifying various medical conditions. As an example, we have chosen three common medical datasets to evalaute the robustness of MedAL. These medical datasets are:
**Messidor Dataset** contains 1200 eye fundus images from 654 diabetic and 546 healthy patients. This dataset was labeled for Diabetic Retinopathy (DR) grading and for Risk of Macular Edema. In this work, we are using Messidor to classify eye fundus images as healthy (DR grade = 0) or as having DR (DR grade > 0). We used 768 images as the combined oracle and training sets, 192 for testing and the remaining 240 as a validation set where we tuned our parameters.
**Breast Cancer Diagnosis Dataset** was part of ICIAR 2018 Grand Challenge [19]. It consists of 400 high resolution images of breast tissue cells belonging to four different classes: Normal, Benign, in-situ carcinoma and invasive carcinoma (having 100 images per class). The dataset was split up into 320 images for the combined oracle and training set and the remaining 80 images for the test set.
**Skin Cancer dataset** contains 900 benign and malignant cell tissue images [20]. We used 700 images as the combined training and oracle sets and 200 images for testing. The class distribution of this dataset was highly biased towards the negative (benign) class (80%) vs positive (malignant) class (20%). To balance the dataset, in each iteration of the AL algorithm, we augmented the positive class samples of the training set using techniques such as cropping, flipping, translating, etc.

Examples from each of these three datasets are shown in Figure 3.

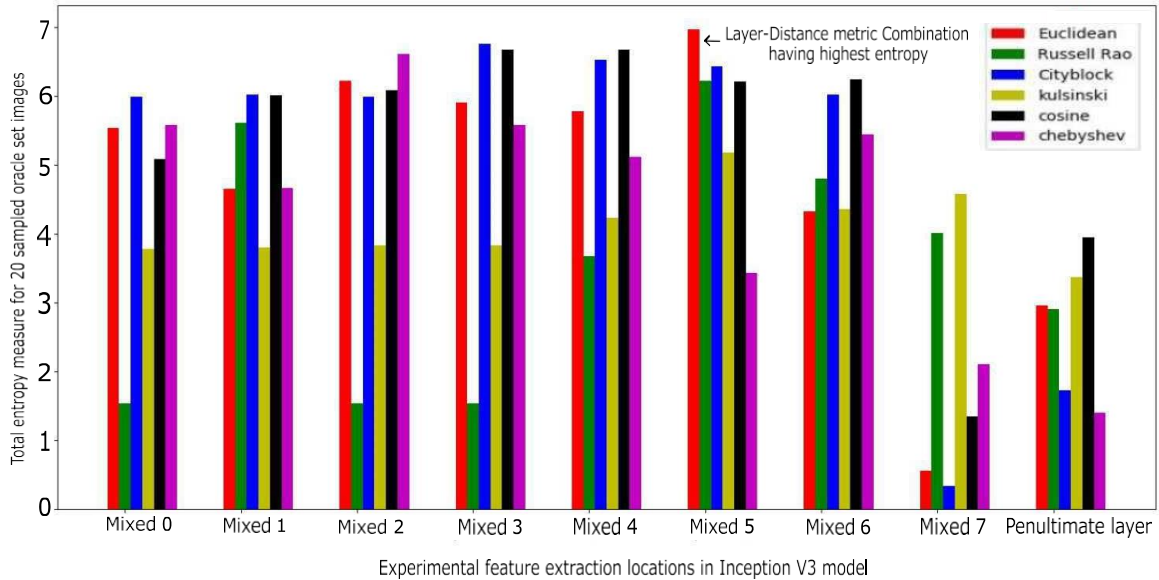

Fig. 4. **Selection criteria for model layer and distance function.** We compare the sum of the predictive entropy computed from 20 oracle set images using various combinations of distance function and feature extraction locations in the Inception V3 network. We find that using the euclidean distance on features extracted from the 'Mixed 5' layer maximizes the predictive entropy.

### B. Implementation Details

We used a Inception V3 [21] network pre-trained on ImageNet to both extract features and classify the images in the AL process. The last layer of the Inception V3 was removed and a Global Average Pooling layer was added, followed by a Fully-Connected layer, to account for different input image resolutions and number of output classes. We used an Adam [22] optimizer with learning rate of $2e-4$ and we kept the default recommended values for $\beta_1$ (0.9) and $\beta_2$ (0.999).

At each AL interation, the model is trained until obtaining 100% accuracy on the training set. Therefore, Messidor's validation set was only used for hyper-parameter tuning and then the same parameters were used in the remaining datasets without the need for a labeled validation set. This is important since, in an AL context, we want to minimize the amount of required labeled data. Furthermore, the model's parameters were reset after each iteration: the standard Inception V3 layers were reset to the pre-trained weights from ImageNet while the new Fully-Connected layer were initialized with random weights using the glorot method [23].

All images were resized to 512x512 pixels and standard dataset augmentation was used during training. Further implementation details for the various datasets are shown in Table I above.

### C. Distance Function Evaluation

To optimize the information gain in our feature extraction approach, we evaluated a set of common distance functions to be used in Eq. 1. We considered 'Euclidean', 'Russellrao', 'City Block', 'Kulsinski', 'Cosine', and 'Chebyshev' as the distance functions [24]. By considering the features extracted from all Inception V3 layers, we can choose the layer which gives us the most information in its feature vector representation.

We trained the model on Messidor's initial training set until it achieves 100% accuracy and, then, we sample 20 images from the oracle set based on each of the previously mentioned distance functions. We repeat this process for all different layers in the Inception V3 network and then we finally compute the entropy of each of the selected oracle set images.

As shown in Figure 4, the distance function that achieves a higher entropy value is the euclidean distance when features are extracted from the Mixed5 layer. Therefore, this was the layer-distance combination that we used in all following experiments.

### D. ORB-based Initial Dataset Formation Evaluation

By considering the dissimilarity of the images to form the initial dataset, our ORB-based initial dataset formation prevents the addition of redundant samples. To validate our ORB-based initial dataset formation approach, we have compared the accuracy using our approach against an approach which forms the dataset via random sampling.

We trained a model until it obtains 100% accuracy on the initial training set in order to compare random sampling and our approach. As presented in Table II our ORB-based approach significantly outperform random sampling in all three datasets. Specifically, our approach results in up to 7.34 percentage points improvement in accuracy over random sampling. The consistent performance improvement across various datasets shows that by capturing the most distinct images from the available unlabeled dataset, we can form a more informative initial training set and potentially reducing the labeling effort from human annotators.

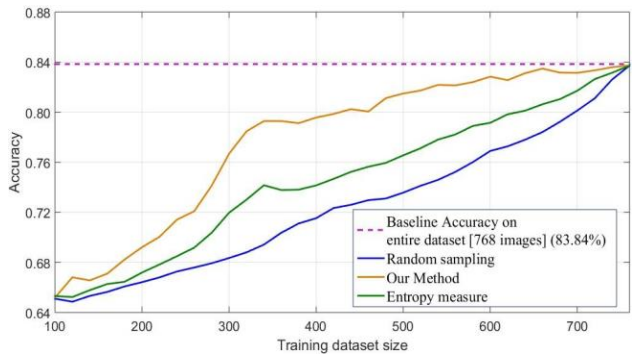 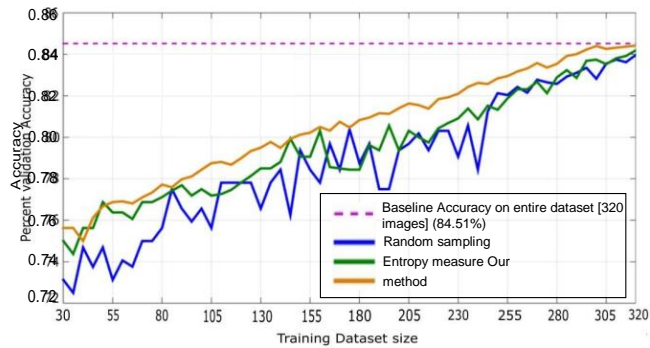

Fig. 5. **Messidor Results.** Test set accuracy as a function of the number of labeled training images. We can see that our method achieves 80% accuracy with 200 less labeled images in the training set than uncertainty sampling. Using our method, it is possible to obtain comparable results to a model trained on the full dataset of 768 images using only 650 labeled images.

Fig. 6. **Breast Cancer Results.** Our method samples images that consistently improve the test set performance. Our method still slightly outperforms both uncertainty and random sampling.

### E. Distance-based Sampling Method Evaluation

To validate our sampling method, we performed a set of experiments and compare its performance to the performance of uncertainty sampling and random sampling techniques on the three datasets mentioned above. These three datasets have different properties: we are using Messidor in a binary classification problem, the Breast Cancer dataset in a multi-class classification problem, while the Skin Cancer dataset is unbalanced.

We monitor the test accuracy of our model after each AL iteration. As shown in Figure 5, our method clearly outperforms uncertainty sampling and random sampling. For instance, with our method we obtain 80% accuracy with 425 images, whereas uncertainty sampling requires 625 images and random sampling requires 700 to achieve the same 80% accuracy. Moreover, by using our method we obtain comparable results to the baseline accuracy using only 650 training images out of 768.

Our method is also consistently better than competing methods in the Breast cancer dataset, as shown in Figure 6, although the difference is not as visible as in Messidor. Our approach reaches 82% accuracy with 230 images in the training set, whereas the uncertainty sampling method requires 250 images and random sampling requires 255 samples to achieve the same accuracy.

Finally, our approach also reaches 69% accuracy on the Skin Cancer dataset while being trained on 460 images, as shown in Figure 7, whereas uncertainty sampling requires 570 training images and random sampling requires 640 examples to achieve the same results. Furthermore, our method achieves 71% accuracy after being trained with 610 images, the same performance as when trained with the full dataset of 700 images.

These results show that our sampling technique works better than both uncertainty sampling and random sampling techniques. The reason is that our approach not only accounts for the degree of uncertainty of the samples but also their feature representations.

## V. DISCUSSION AND FUTURE WORK

As shown in the evaluation results above, our sampling technique outperforms both uncertainty sampling and random sampling techniques for AL in the medical image domain. For each of the three datasets, our approach obtains a higher overall accuracy using fewer training examples. These results support the underlying assumptions of our approach: that the most informative examples are the ones where the model has the highest uncertainty and greatest distance/dissimilarity to the training examples.

We can also see that our method achieves comparable results to the baseline method using fewer labeled images. This may suggest that there are some images that are not informative and our method safely discards them. Moreover, in Figure 5 we can see that the model's performance improves almost linearly using random sampling while, when using our method, the improvement of the model is more dramatic in the early stages of the sampling process. This may suggest that there are a small number of very informative images present in the oracle set that our method selects first and that, if we had access to a larger unlabeled dataset, this rapid performance improvement could continue.

In the future, we intend to apply this technique to help in the creation of a new private dataset for Diabetic Retinopathy grading. We have a large set of unlabeled eye fundus images and we want to test wether this method scales for larger dataset sizes. Finally, we are also collecting the time a medical doctor takes to label each image and we want to study whether the labeling time correlates with our AL sampling method selection. In other words, we want to check if our method selects the images that the medical doctors find more challenging.

## VI. CONCLUSION

In this paper we introduce MedAL, a novel AL sampling method that queries the unlabeled examples that maximize the average distance to all training set examples in a learned feature space. Furthermore, we show that we can use this sampling method to create an initial training set that leads to better generalization when compared with a randomly sampled dataset.

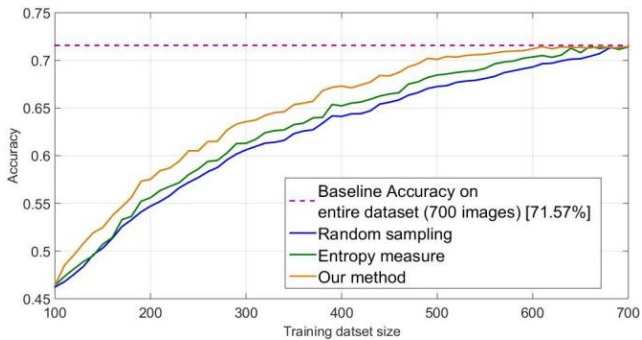

Fig. 7. **Skin Cancer Results.** Our method also works on an unbalanced dataset. Using our sampling technique, we achieve the baseline results using 610 labeled images, 90 images less than the full dataset.

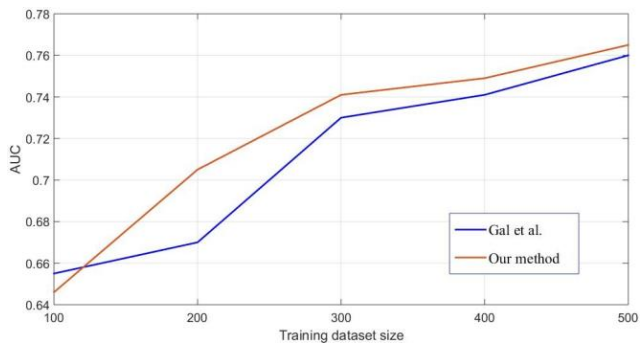

Fig. 8. **AUC on the Skin Cancer dataset.** Our sampling method slightly outperforms Gal *et al.*'s [25] deep bayesian AL method.

We evaluated our method on 3 medical imaging datasets with different properties: Messidor, Breast cancer dataset and Skin cancer dataset. We show that MedAL is robust, performing well on both binary and multi-class classification problems, and also on balanced and unbalanced datasets. Our experiments show that we can obtain the same results as a model trained on the entire dataset using less labeled examples.

For instance, with MedAL we were able to train a model that achieves comparable results to a model trained on the entire 768 image dataset using only 650 images. Moreover, MedAL performs better than uncertainty and random sampling. Our approach reaches 80% accuracy using just 425 images which corresponds to 32% and 40% reduction compared to uncertainty and random sampling methods respectively. In conclusion, MedAL achieving consistantly better performance than all competing methods on all 3 medical imaging datasets.

ACKNOWLEDGMENT

This work was supported in part by the European Regional Development Fund through the Operational Programme for Competitiveness and Internationalisation–COMPETE 2020 Programme and in part by the National Funds through the Fundação para a Ciência e a Tecnologia within under Project CMUPERI/TIC/0028/2014.